\documentclass{isprs}
\usepackage{multirow}
\usepackage{subfigure}
\usepackage{setspace}
\usepackage{geometry}
\usepackage{epstopdf}
\usepackage[labelsep=period]{caption}
\usepackage[british]{babel} 
\usepackage[hang]{footmisc}

\usepackage{booktabs}
\usepackage{threeparttable} 
\usepackage{amsmath}

\usepackage[authoryear]{natbib}
\usepackage[hidelinks]{hyperref}
\usepackage{multirow} 
\usepackage{tabularx}

\usepackage{listings}
\usepackage{xcolor}

\begin{document}

\title{Zero-Shot Building Age Classification from Facade Image Using GPT-4}
\date{Dec 2023}

\author{Z. Zeng\textsuperscript{1}, J. M. Goo\textsuperscript{1}, X. Wang\textsuperscript{1}, B. Chi\textsuperscript{2}, M. Wang\textsuperscript{1}, J. Boehm\textsuperscript{1}\thanks{Corresponding author}}

\address{
	\textsuperscript{1 }Department of Civil, Environmental and Geomatic Engineering, University College London, Gower Street, London, WC1E 6BT UK\\
	\textsuperscript{2 }Department of Geography, University College London, Gower Street, London, WC1E 6BT UK \cr – \{zichao.zeng.21, june.goo.21, xinglei.wang.21, bin.chi, meihui.wang.20, j.boehm\}@ucl.ac.uk\\
}

\commission{XX, }{YY}
\workinggroup{XX/YY}
\icwg{}

\abstract{
A building's age of construction is crucial for supporting many geospatial applications. Much current research focuses on estimating building age from facade images using deep learning. However, building an accurate deep learning model requires a considerable amount of labelled training data, and the trained models often have geographical constraints. Recently, large pre-trained vision language models (VLMs) such as GPT-4 Vision, which demonstrate significant generalisation capabilities, have emerged as potential training-free tools for dealing with specific vision tasks, but their applicability and reliability for building information remain unexplored. In this study, a zero-shot building age classifier for facade images is developed using prompts that include logical instructions. Taking London as a test case, we introduce a new dataset, FI-London, comprising facade images and building age epochs. Although the training-free classifier achieved a modest accuracy of 39.69\%, the mean absolute error of 0.85 decades indicates that the model can predict building age epochs successfully albeit with a small bias. The ensuing discussion reveals that the classifier struggles to predict the age of very old buildings and is challenged by fine-grained predictions within 2 decades. Overall, the classifier utilising GPT-4 Vision is capable of predicting the rough age epoch of a building from a single facade image without any training. The code and dataset are available at \href{https://zichaozeng.github.io/ba_classifier}{https://zichaozeng.github.io/ba\_classifier}.
}


\maketitle

\sloppy
\section{INTRODUCTION}\label{sec:INTRODUCTION}
Individual building age is one of the significant attribute data for building information modelling and urban digital twins. For the building itself, energy demand and housing price are linked to building age \citep{aksoezen2015building, law2019take}. In general, building age has considerable impact on historical architecture preservation, urban planning and disaster management \citep{SUN2022103787, 10018300}. However, due to numerous missing data or irregular data collection methods, there are innumerable buildings for which we do not know their age. 

Within the last decade, many scholars were attempting to establish building age using deep learning. \citet{Li2018Estimating} presented a novel method for estimating building age from Google Street View images using a convolutional neural network (CNN) for image features extraction and support vector machine for construction year regression. \citet{10.1145/3206025.3206060} introduced the automated age estimation method for unconstrained facade photographs from patch-level visual feature learning to global age classification. Afterwards, \citet{DESPOTOVIC201929}, \citet{sun2021automatic}, and \citet{10018300} continually explored possible building age estimation methods using deep learning techniques from facade views.

With the emergence of deep learning models, the demand for training data has grown significantly. Due to the large regional differences in building architecture, transfer of models between regions is uncertain and the number of training images for any specific region is typically limited. Today's foundation models are often trained on data sets with 100s of millions of images, see for example JFT \citep{hinton2015distilling} or CLIP \citep{radford2021learning}. We therefore need to consider for each application if training a new model using a new labelled dataset is efforts well invested, or if using a foundation model performs just as good. We explore this idea in the context of estimating the age of a building form a single image of its facade. We do this by applying a foundation model without additional training.

\section{RELATED WORK}\label{sec:LR}
\subsection{Building Age Epoch}
\begin{figure*}[t]
\begin{center}
		\includegraphics[width=1\linewidth]{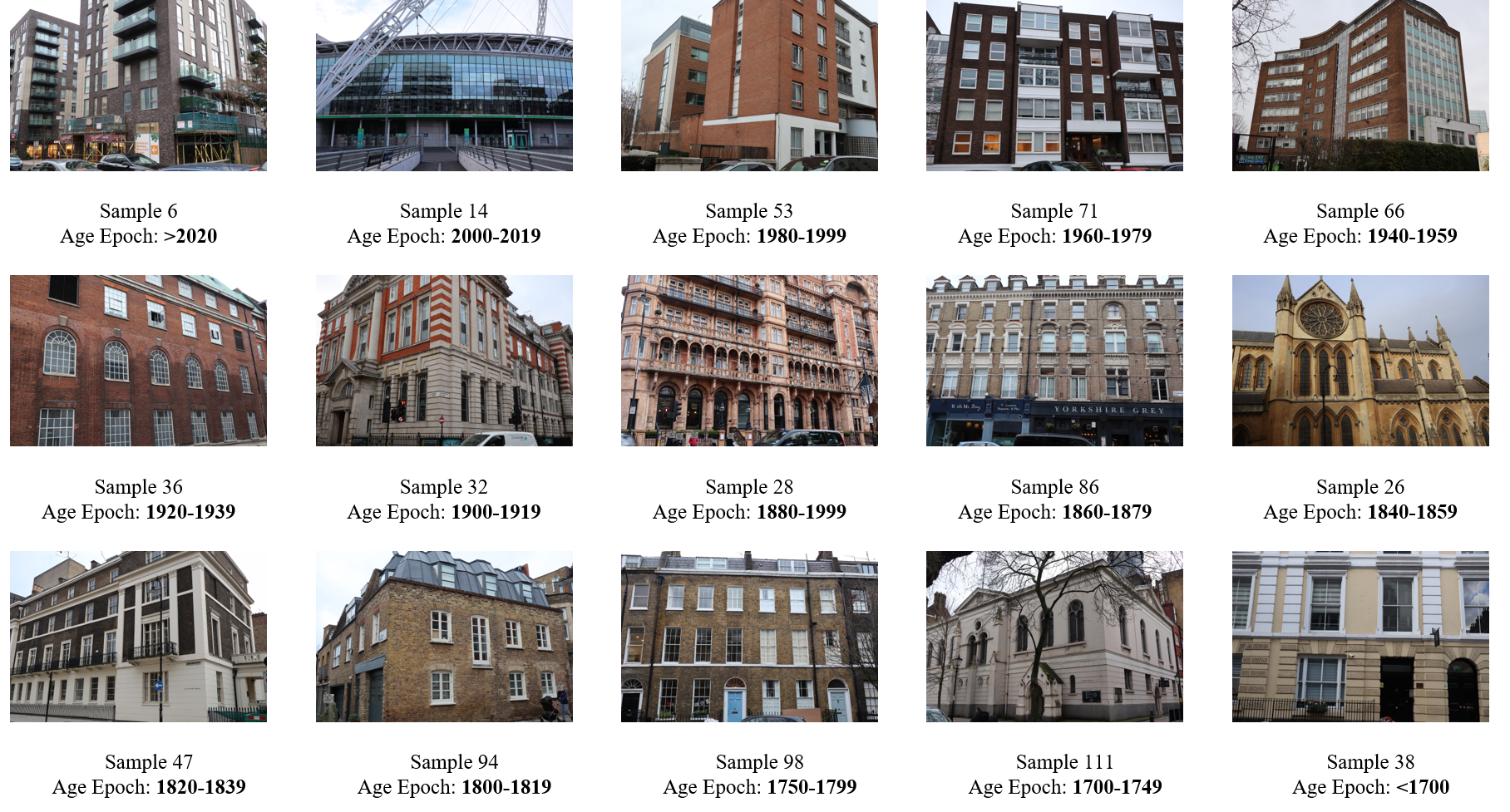}
	\caption{Sample Images contained in  FI-London with age epoch}
   \label{fig:SampleDistribution}
\end{center}
\end{figure*}
The age of building stock in our built environment is a crucial information for estimating energy demand, urban planning, cultural heritage protection, disaster resilience, etc. Building age is a crucial indicator in energy consumption analysis and is an important variable for energy demand estimation \citep{aksoezen2015building, GARBASEVSCHI2021101637}. During the last few decades, a considerable number of studies on green or energy-efficient buildings, have shown that the age of construction is highly related to energy use and has thus become a key indicator to define sustainable construction \citep{aksoezen2015building}. This has also led to building age being closely related to housing prices, in the real estate industry and in urban planning \citep{tam1999feng, stanley2016price, law2019take}.  In addition to impacts on energy analysis, the assessment of potential disasters such as earthquake and tsunami damages utilise the year of construction as a parameter in damage fragility curve models \citep{nagao2011analysis, del2017empirical}. Besides, the estimation of the building age is essential for the preservation of historic buildings and cultural heritage, which can help to identify which buildings need to be preserved or restored, especially in urban planning and development.

Due to the remarkable success of computer vision on particular applications such as medicine and engineering, images of buildings have been considered as a source for estimate building attributes, which can fill gaps in the available data effectively and directly. However, most techniques in the study of architectural imagery concentrate on categorising structures according to distinct styles of architecture \citep{DESPOTOVIC201929}. \citet{6247857} developed a method for classifying architectural styles in images using an irregular rectangular lattice and a combination of pixelwise classifiers and object detectors to analyse and segment building facades. Similarly, \citet{shalunts2011architectural} categorised different construction styles by scale invariant feature transform (SIFT) from facade window components. \citet{law2019take} , on the other hand, built a pipeline based on deep convolutional neural network (CNNs) for extracting visual features from satellite and street view images and combines them with pre-existing building attributes (e.g., age, size, and accessibility) to estimate house prices in London, UK.

In the classification task of building age epochs, \citet{10.1145/3206025.3206060} proposed a framework based on CNNs to predict building age. The method aggregates patch-level predictions to derive a global estimate for the entire building. \citet{DESPOTOVIC201929} grounded this framework to infer the year of construction and then analyse the heating energy demand. \citet{10018300} combined geographic information system (GIS) data to predict the age of buildings from street view imagery. In this study, they compared networks based on deep CNNs with a ViT-based architecture and found that the ViT-based Swin transformer improved the classification accuracy effectively.

\subsection{Vision Language Model}
Pre-trained Large Language Models (LLMs), especially ChatGPT by \citet{openai2023gpt4}, have not only achieved extreme performance in natural language processing (NLP) tasks with zero-shot setting \citep{qin2023chatgpt}, but have also inspired new attempts in other fields. In geospatial science, \citet{roberts2023gpt4geo}, \citet{li2023autonomous} and \citet{wang2023would} indicated GPT's significant capabilities in geographic knowledge and reasoning including human mobility prediction, country outlines creation, travel routes planning, supply chain analysis, etc. Notably, these case studies are zero-shot learning examples using LLMs without any training and tuning, which means no labelled training data is needed to complete prediction tasks intelligently. 


In addition to textual tasks, pre-trained models that understand both images and language, often called Vision Language Models (VLMs) or multi-modal models, are increasingly used in image understanding tasks.
Various large VLMs like CLIP \citep{radford2021learning}, BLIP \citep{li2022blip}, GLIP \citep{li2022grounded}, and Grouding-DINO \citep{liu2023grounding} have demonstrated their abilities in visual understanding and have attained a high accuracy with zero-shot learning in visual benchmark tasks such as classification, segmentation, detection, and depth estimation. These models are trained on a significant number of image-text pairs and are not limited to pre-defined classes. This allows large VLMs to identify unlearned object from images by an operator's text input referred to as a prompt \citep{radford2021learning}. 

Besides being able to identify objects or scenes without prior specific training, VLMs can be used effectively across different geographic locations, thanks to their integration with LLMs that include training data with large amounts of geographic information. Traditionally, models trained in one specific area often struggle when used in a different location \citep{kedron2022replication}. However, VLMs that incorporate language understanding can easily adapt and be applied in various geographic areas. This adaptability is due to their training on very large and diverse data, making it easier to bridge geographical differences \citep{zhang2022migratable}.

\begin{figure*}[t]
\begin{lstlisting}[language=Python]
{
    "type": "text",
    "text": """
        Your task is to predict the age epoch of a building in London based on the image provided by users.
        
        You will be presented with <building>, an image containing a main building. You need to infer the most likely <building_age_epoch>.

        Only select <building_age_epoch> from this list: [">2020", "2000-2019", "1980-1999", "1960-1979", "1940-1959", "1920-1939", "1900-1919", "1880-1899", "1860-1879", "1840-1859", "1820-1839", "1800-1819", "1750-1799", "1700-1749", "<1700"].

        Organize your answer in the following format containing two keys: 
        {
            "age": <building_age_epoch>,
            "reason": ""
        }

        The meaning of two keys:
        - "age": the most likely <building_age_epoch> chosen from the provided list.
        - "reason": a concise explanation supporting your prediction. Please do not use line breaks in the reason.
    """
}
\end{lstlisting}
\caption{The prompt for building age classification used in GPT-4 Vision\label{fig:prompt}}
\end{figure*}

\begin{figure}[t]
\begin{center}
		\includegraphics[width=1\columnwidth]{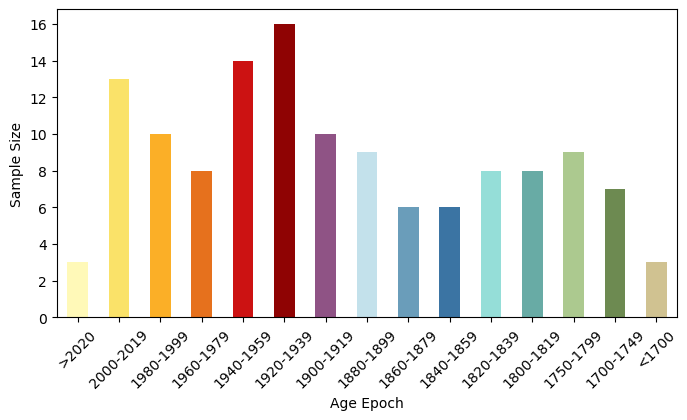}
	\caption{Distribution of Building Epochs in FI-London}
    \label{fig:SampleImages}
\end{center}
\end{figure}

For architectural vision tasks such as building age epoch classification, most datasets have restricted geographic locations and small sample size. Large VLMs like GPT-4 \citep{openai2023gpt4} emerge as new contenders to classify or predicate structural attributes without any training data. Therefore, in this study, we explore the zero-shot ability of GPT-4 as a classical large VLM for the building age classification task.



\section{DATA AND EXPERIMENT}\label{sec:METHOD}

\subsection{Dataset}
A related work by \citet{DESPOTOVIC201929} created a closed dataset that was obtained by web scraping. The dataset contains a total of 3865 facade images of 2065 buildings. The buildings from this dataset are all individual houses, which present a more homogeneous architectural style. The chronological classification of this dataset covers modern buildings from 1969 to 2010, and this time span makes the chronological differentiation of building facade features less obvious. In another related work, \citet{sun2021automatic} combined Google Street View data with a building age dataset from Amsterdam, aiming for a more detailed analysis. 

For the purpose of this study, a dataset of facade images containing building-specific attributes of London (referred to as FI-London) was created. Focusing on Brent and Camden in London, FI-London contains a total of 131 high-resolution building facade images (6000 $\times$ 4000 pixels). These images contain individual building facades of varying building types such as residential apartment blocks, terraced houses, commercial properties, etc. Furthermore, the images contain occlusions such as pedestrians, cars and scaffolding. This is a potential distraction, but a relevant challenge for GPT-4 Vision. FI-London covers 15 different architectural age epochs seen in Figure \ref{fig:SampleDistribution}, derived from the Colouring Cities project by \citep{hudson2019colouring, colouring_london_github}. Due to the unbalanced sample distribution seen in Figure \ref{fig:SampleImages} and small sample size, FI-London can currently only be used for testing. We chose London as a case study location because the city's buildings vary greatly in age epoch and facade style \citep{jones2005architecture}. Many of the historic buildings are relatively well preserved, providing an ideal test environment for evaluating GPT-4 Vision.



\subsection{Building Age Classier}

\begin{figure}[t]
\begin{center}
		\includegraphics[width=0.8\columnwidth]{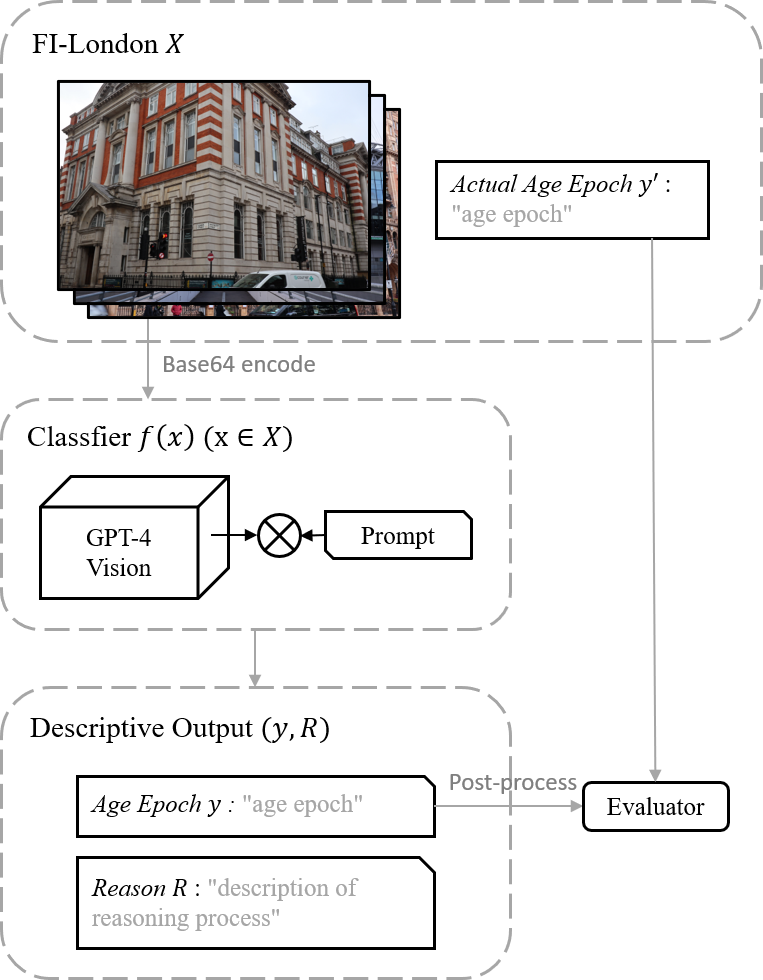}
	\caption{Framework for zero-shot classification of age epoch}
    \label{fig:framework}
\end{center}
\end{figure}

We propose a training-free classifier (zero-shot) \(f\) to identify the age epochs of buildings using GPT-4 Vision, and the detailed process is shown in Equation \ref{equ:model} and Figure \ref{fig:framework}. Firstly, we encode the input image \(x\) in FI-London \(X\) into base64 format, and the corresponding building age epoch - i.e. ground truth (\(y'\)) - is stored as a JSON file for evaluation purposes. We craft a series of command prompts to perform zero-sample classification directly using GPT-4 Vision. Finally, the predicted age epoch (\(y\)) and descriptive reasoning process (\(R\)) are output by the classifier (\(f\)) with prompts. In the prompts, we revealed to GPT-4 that the input image \(X\) is in London. This is because the task of this study is to predict the age epoch, not the specific geographic location. 


\begin{equation}\label{equ:model}
(y, R) = f(x)(x \in X)
\end{equation}

\begin{tabularx}{\columnwidth}{lX}
where & $y$ = Predicted Age Epoch\\
& $R$ = Descriptive Reasoning Process\\
& $f$ = GPT-4 Vision with Prompt\\
& $x$ = Input Image in Test Set $X$
\end{tabularx}

The user's prompt has major influence in LLM-based tasks. Seen in Figure \ref{fig:prompt}, the instruction prompt consists of four components. For a start, the instruction prompt includes a general task description for initialising the GPT-4 Vision. Next, the encoded input image \(x\) and the expected age epoch \(y\) are defined as $<$building$>$ and $<$building\_age\_epoch$>$, respectively, by which a context is generated to improve GPT-4 Vision's understanding of the task and data. Furthermore, since building age estimation in this study is regarded as a categorisation task, we provide a list of building age epochs as a way to constrain the output of GPT-4 Vision. Finally, given that GPT-4 Vision is still unable to directly output JSON-formatted results at the time of our experiments, we have also included instructions for formatting the results within the prompt. In addition to the age epochs \(y\), we require the output of the reasoning process \(R\), which not only validates the  thinking process of GPT-4 Vision, but also supports the subsequent discussion.



\subsection{Evaluation Metrics}

We employ standard evaluation metrics in the field of image classification to measure the performance of our model. Specifically, \textbf{Precision} and \textbf{Recall} are used to evaluate the model's performance on individual age epoch. Meanwhile, we use \textbf{Micro F1-score} as a composite metric to evaluate the overall performance on the whole multi-categorisation task, whose formula is given in Equation \ref{equ:micro_F1}. It is worth noting that \textbf{Micro F1-score} is the same as \textbf{Accuracy} in multi-categorisation task.
\begin{equation}
\label{equ:micro_F1}
\begin{split}
    \text{Micro F1} &= 2 \times \frac{\text{Precision}_\text{Micro} \times \text{Recall}_\text{Micro}}{\text{Precision}_\text{Micro} + \text{Recall}_\text{Micro}} \\
    &= \frac{ \sum TP }{ \sum TP + \sum FP + \sum FN} \\
    &= \frac{ \sum TP }{\text{Total number of instances}} = \text{Accuracy}
\end{split}
\end{equation}

\begin{tabularx}{\columnwidth}{lX}
where & \textbf{Micro Precision} = the ratio of the total number of correct predictions across all categories to the total number of predictions made across all categories. \\
& \textbf{Micro Recall} = the ratio of the total number of correct predictions across all categories to the total number of actual positives across all classes. 
\end{tabularx}

Considering that there is often a high degree of similarity in architectural styles between consecutive age epochs, and that we classify age epochs into relatively short time intervals of two decades (5 decades before 1800), the task is rather challenging. This is especially true for buildings that are at the intersection of two epochs, where they can easily be misclassified. For this reason, we introduced the \textbf{Mean Absolute Error (MAE)} of the age epochs as another performance metric (see Equation 3 for details) to quantify the difference between the mid-year of the predicted age epoch \(m_{y}\) and the mid-year of the true age epoch \(m_{y'}\), seen in Equation \ref{equ:MAE}.

\begin{equation}\label{equ:MAE}
\text{MAE} = \frac{1}{N} \sum_{i=1}^{N} |(m_{y_i}-m_{y'_i})/10|
\end{equation}
\begin{tabularx}{\columnwidth}{lX}
where & $m_{y_i}$ = Mid Year of Predicted Age Epoch\\
& $m_{y'_i}$ = Mid Year of Actual Age Epoch\\
& $N$ = Instance Size
\end{tabularx}

In addition, we perform an intuitive analysis of the results by constructing  \textbf{Confusion Matrices}. Besides a normal confusion matrix, we design specific confusion matrices in which those cases that were incorrectly classified as adjacent ages (i.e., one wrong epoch or two wrong epochs) were also considered as correct predictions. These specific confusion matrices approach allow us to intuitively analyse the degree of chronological prediction of GPT-4 Vision.






       

\subsection{Experiment Setting}
One of the advantages of our proposed GPT-4-based classifier is that it does not require any training and the inference process is conducted through the API provided by OpenAI, so it does not require significant computational resources. The experiments were performed on an Ubuntu Server with an Intel(R) Core(TM) i5-13600KF @ 3.50GHz, 64GB of random-access memory, and a GeForce RTX 4070 Ti with 12GB of graphic memory. The model we used is \emph{GPT-4 Vison Preview}. The inference speed was roughly 10 seconds per image, and the cost was \$2.08 for 131 images.
\section{RESULT AND DISCUSSION}\label{sec:RESULT}

\begin{table}[b]
\centering

\resizebox{0.75\linewidth}{!}{
\begin{threeparttable} 
\begin{tabular}{c|ccc|c}
\toprule[2pt]
\multirow{2}{*}{Age Epoch}&Precision&Recall&F1&MAE \\
&\%&\%&\%&10 yrs\\
\midrule
$<$1700&0.00&0.00&0.00&16.95\tnote{1}\\
1700-1749&33.33&14.29&20.00&7.21\tnote{2}\\
1750-1799&42.86&33.33&37.50&2.33\tnote{2}\\
1800-1819&25.00&62.50&35.71&0.44\\
1820-1839&40.00&25.00&30.77&1.06\\
1840-1859&0.00&0.00&0.00&2.07\\
1860-1879&25.00&16.67&20.00&2.33\\
1880-1899&35.29&60.00&44.44&0.40\\
1900-1919&25.00&20.00&22.22&2.00\\
1920-1939&56.25&56.25&56.25&0.75\\
1940-1959&100.00&21.43&35.29&1.57\\
1960-1979&28.57&75.00&41.38&0.50\\
1980-1999&50.00&40.00&44.44&0.0\\
2000-2019&55.56&76.92&64.52&2.00\\
$>$2020&0.00&0.00&0.00&1.00\tnote{1}\\
\midrule
Total&\multicolumn{2}{c|}{Accuracy/Micro F1}&39.69&0.85 \\
\bottomrule[2pt]
\end{tabular}
\begin{tablenotes}
\footnotesize
\item[1] Mid-year \(m_{y'_i}\) is set to 1700/2020.         
\item[2] 5-decade interval age epochs.     
\end{tablenotes} 
\end{threeparttable}
}
\caption{Performance of zero-shot classifier on FI-London}
\label{tab:perfomance}
\end{table}

\subsection{Experimental Result}

\begin{figure*}[t]
\begin{lstlisting}[language=Python]
{
    "age": "1940-1959", # ground truth: 1940-1959
    "reason": "The building's architectural style suggests a post-war design common between the 1940s and 1950s in London, characterized by the streamlined, functional appearance and the use of stone cladding.", 
    "ID": 55
} # Correct Result,
{
    "age": "<1700", # ground truth: 1840-1859
    "reason": "The building in the image has Gothic architectural features, such as pointed arch windows, flying buttresses, and a large rose window, which are typical of medieval European churches. Given that Gothic architecture was most prevalent from the 12th to the 16th centuries, the building likely predates the 1700s.", 
    "ID": 26
} # Incorrect Result,
{
    "age": "1973-1997", # ground truth: 1960-1979
    "reason": "The building in the image appears to be the British Library in London, which was constructed between 1973 and 1997. The architectural style is indicative of the late 20th century, with its large, blocky form, red brickwork, and lack of ornamentation typical of the Brutalist style which was popular from the 1950s to the mid-1970s but with construction periods extending into the 1980s and 1990s for some Brutalist buildings.", 
    "ID": 39
} # Hallucination Result,
\end{lstlisting}
\caption{Output example of a correct result, a incorrect result and a "hallucination" result}
\label{fig:result}
\end{figure*}

\begin{figure}[t]
\begin{center}
		\includegraphics[width=0.5\columnwidth]{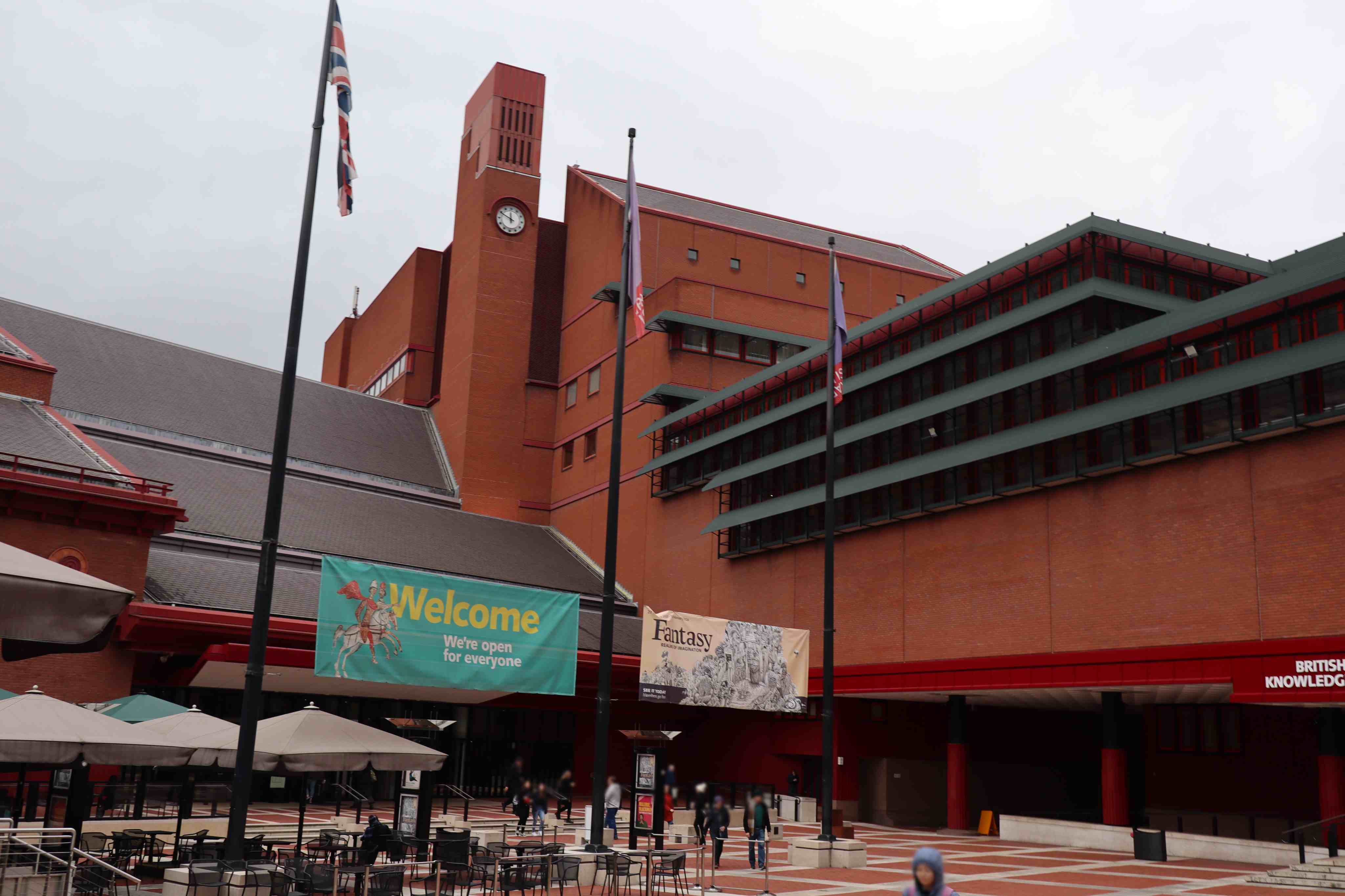}
	\caption{Facade image (of British Library) of the hallucination result: ground truth \(y'_i\) 1960-1979 $\rightarrow$ age epoch \(y_i\) 1973-1997}
    \label{fig:illusion}
\end{center}
\end{figure}


For the 131 building facade images in the FI-London dataset, GPT-4 Vision produced predictions, including 52 correct and 79 incorrect cases, with our designed no-training classifier. Three visual example results are shown in Figure \ref{fig:result}. It is worth noting that among these 79 negative cases, one particular case was identified as a "hallucination", i.e., "the generation of content that strays from factual reality or includes fabricated information" \citep{rawte2023survey}. In this case, GPT-4 fabricates a new age epoch (1973-1997) which is outside the list of categories we postulate within the prompt. However, investigating the input image in Figure \ref{fig:illusion} and analysing the reasoning given by GPT-4 in Figure \ref{fig:result}, we can find that GPT-4 succeeded in recognising the image as the British Library and gave a more accurate prediction. This phenomenon reveals that GPT-4 Vision can rely to some extent on its wider knowledge to make accurate judgements when making predictions, which has different insight from traditional deep learning classifier. Despite its more accurate prediction, we defined it as a "hallucination" result which strays from prescribed categories.

In addition, it also suggests that GPT-4 Vision may be over-reliant on the information in the training dataset, or that the instruction cues we provide need to be further optimised to improve prediction accuracy. Overall, this "hallucination" case provides an interesting perspective on the potential and limitations of GPT-4 Vision in recognising and understanding building facade images.

Through the subsequent comprehensive analysis of the prediction results (shown in Table \ref{tab:perfomance}), we aim to provide insights into the performance of GPT-4 Vision on building age epoch estimation, assess its accuracy and reliability, and discuss its possible implications for future research on architectural style recognition.

\subsection{Performance in Each Age Epoch}
\begin{figure}[t]
    \centering
    \subfigure{
        \includegraphics[width=0.5\columnwidth]{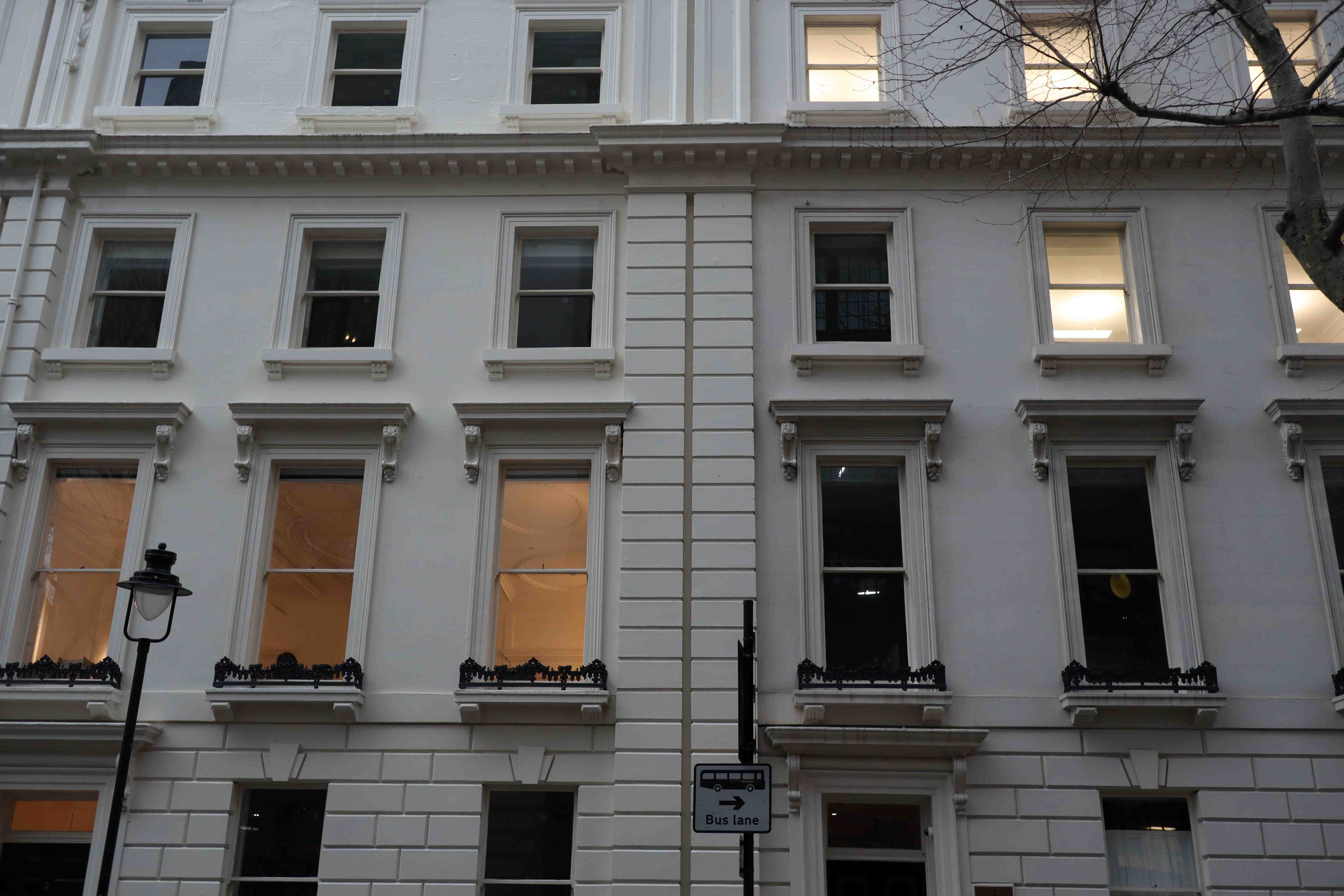}
        }
    \,
    \subfigure{
        \includegraphics[width=0.5\columnwidth]{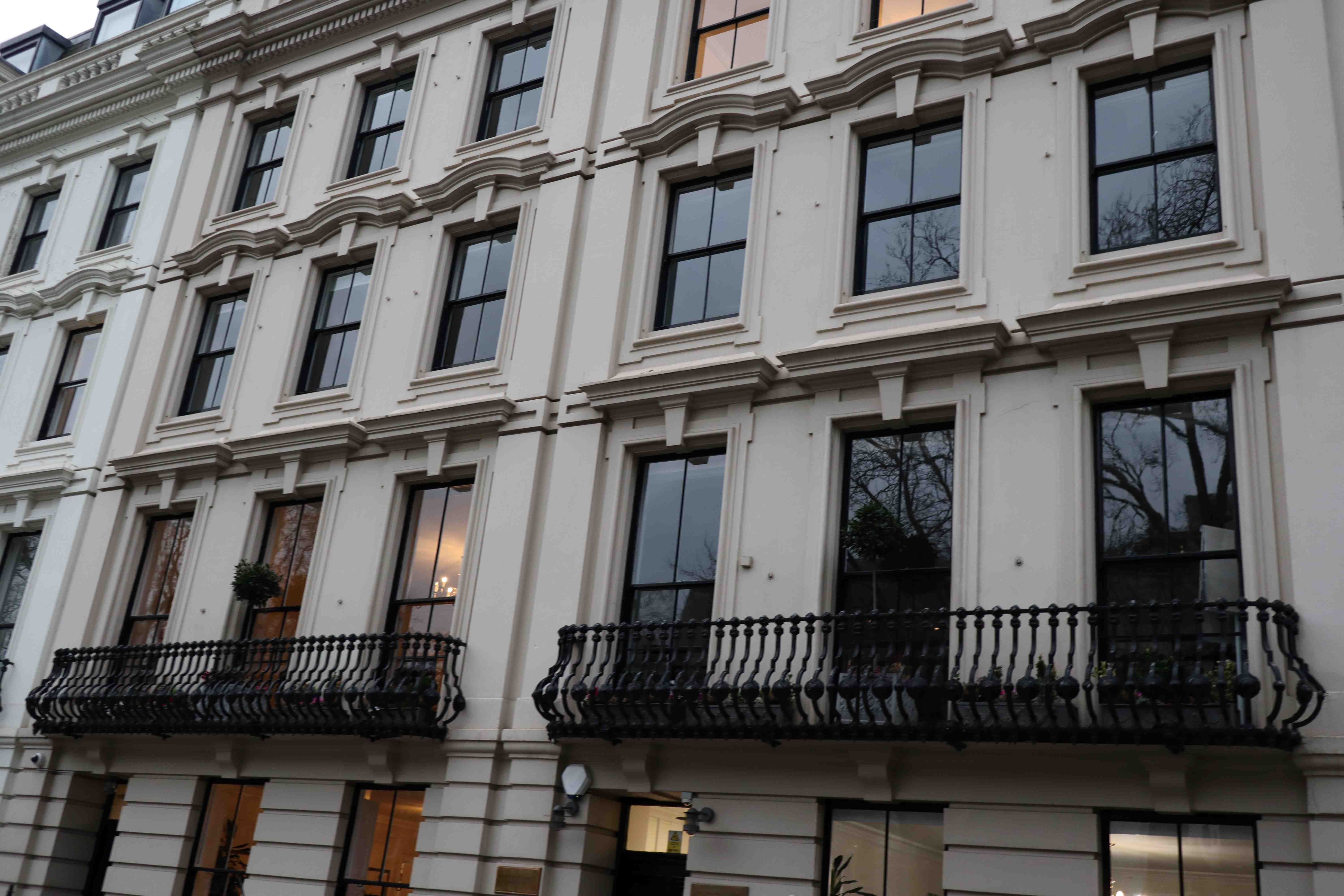}
        }
    \,
    \subfigure{
        \includegraphics[width=0.5\columnwidth]{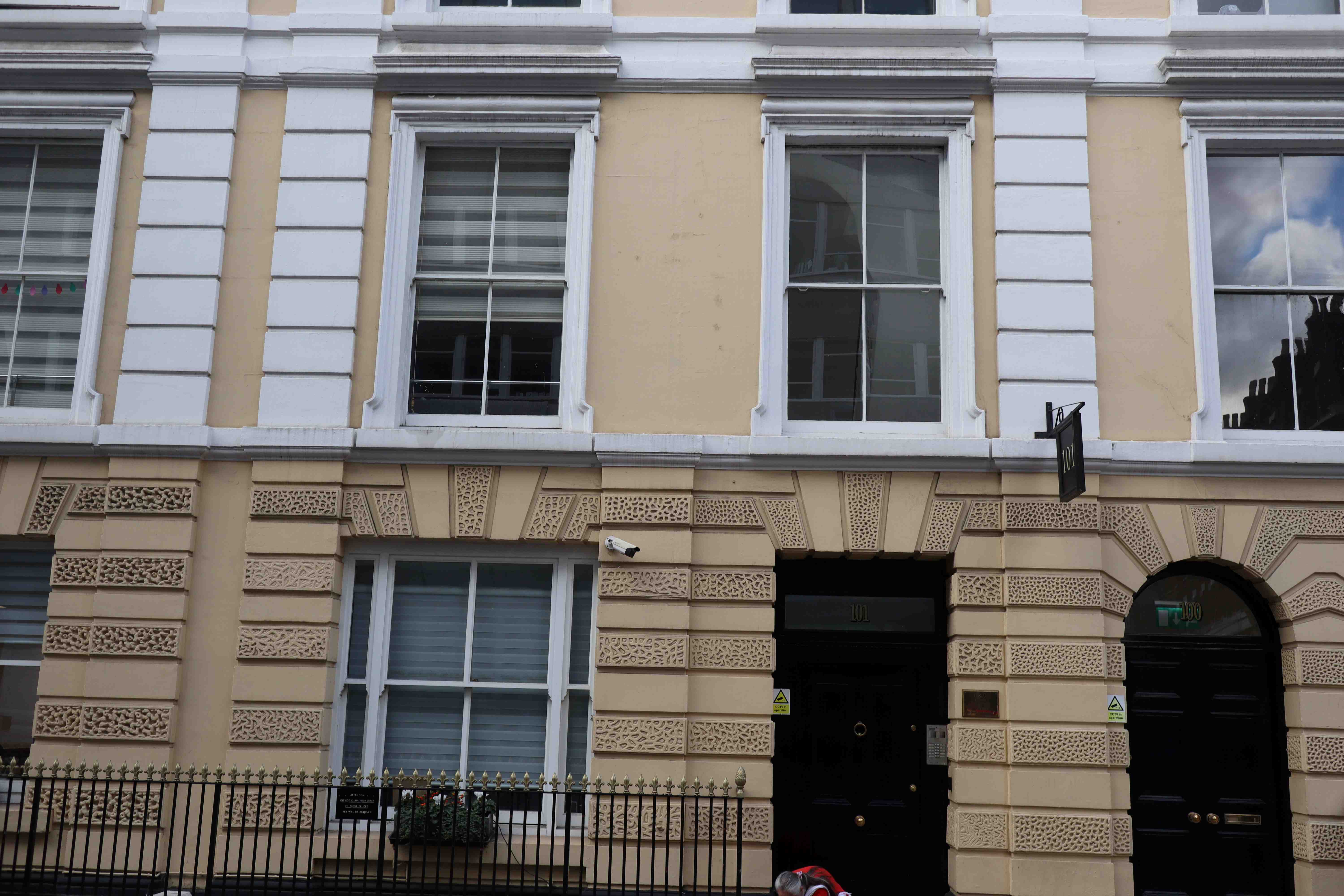}
        }
    \centering
    \caption{Facade images in age epoch $<$1700}
    \label{fig:age_epoch_1700}
\end{figure}

The precision, recall and F1 score of age epochs $<$1700, 1840-1859, and $>$2020 is 0\%, indicating that GPT-4 Vision’s performance for these age epochs is very low, with no correct predictions. This may be due to the insufficient number of building facade images in these age epochs in the testing data, or the architectural features in these age epochs are not obvious. 

GPT-4 Vision, on the other hand, achieves a high performance in age epochs 1920-1939 and 2000-2019. The precision, recall and F1 score of both are over 50\%. Most facade images built in these age epochs are classified accurately. However, all other epochs are not well classified with low F1 scores (20\%-50\%). 

Among them, age epoch 1940-1959 has a high precision (100\%) but a relatively low recall (21.43\%), which means that GPT-4 Vision can perform accurate prediction in this age epoch, but ignores a large number of buildings that are constructed within this age epoch.

In contrast, age epoch 1960-1979 has a high recall (75\%) but a relatively low precision (28.57\%), suggesting that GPT-4 Vision is able to cover most of the buildings that are actually in this age epoch, but at the same time misclassified many samples that do not fall into this age epoch. Similarly, age epoch 1800-1819 has 25\% precision with 62.5\% recall, and age epoch 1880-1899 has 35.29\% precision with 60\% recall.

\subsection{Offset Chronology}

Although GPT-4 Vision performed slightly less well in detailed prediction, by analysing the Mean Absolute Error (MAE) we observed that the bias in the prediction results was not large with 0.85 decade. For most of the age epochs (1750-1799, 1820-1839, 1840-1859, 1860-1879, 1900-1919, 1940-1959, 2020-2019, and $>$2020), the MAEs are only between one and three decades, suggesting that building facade images from these chronological periods tend to be classified into adjacent age epochs. This phenomenon reflects the similarity of architectural styles between these and their adjacent age epochs, as well as in the subtle differences in building facade characteristics and their impact on classification accuracy.

For the age epochs 1800-1819, 1880-1899, 1920-1939, 1960-1979, and 1980-1999, the MAEs are all less than one decade, a result that highlights the high predictive accuracy of GPT-4 Vision within these specific age epochs. This not only suggests that the facade features of buildings in these age epochs are relatively distinct, but also implies that these features are similar to some extent to buildings in the most neighbouring age epochs. In particular, as mentioned above, age epochs 1800-1819, 1880-1899 and 1960-1979 exhibited higher recall and lower precision, while age bands 1920-1939 and 1980-1999 demonstrated higher F1 scores.

However, the prediction difficulty is still high for the age epoch 1700-1749 with a high MAE of 7.21 decades. The reason may be that these old buildings have been renovated or rebuilt in the past time seen in Figure \ref{fig:age_epoch_1700}, which makes the prediction difficult.

\subsection{General Performance}
\begin{figure}[!ht]
    \centering
    \subfigure{
        \includegraphics[width=0.98\columnwidth]{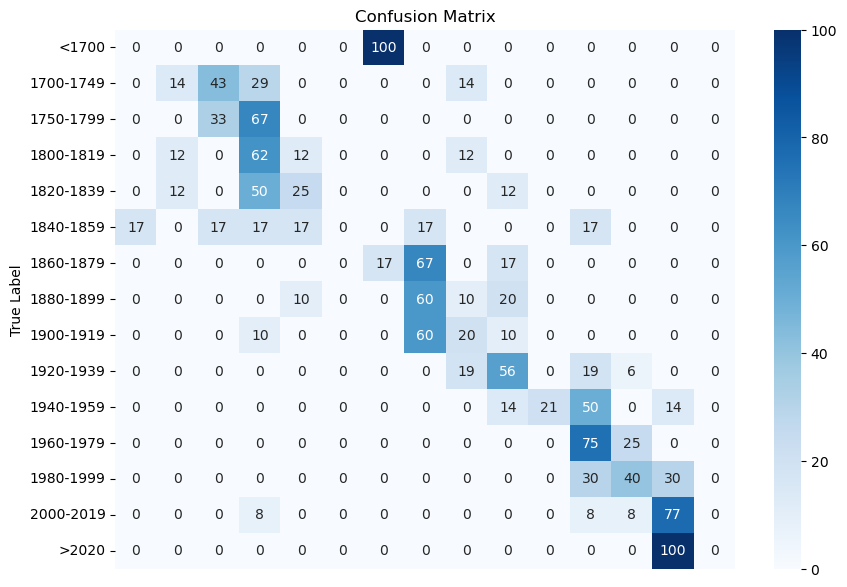}
        }
    \,
    \subfigure{
        \includegraphics[width=0.98\columnwidth]{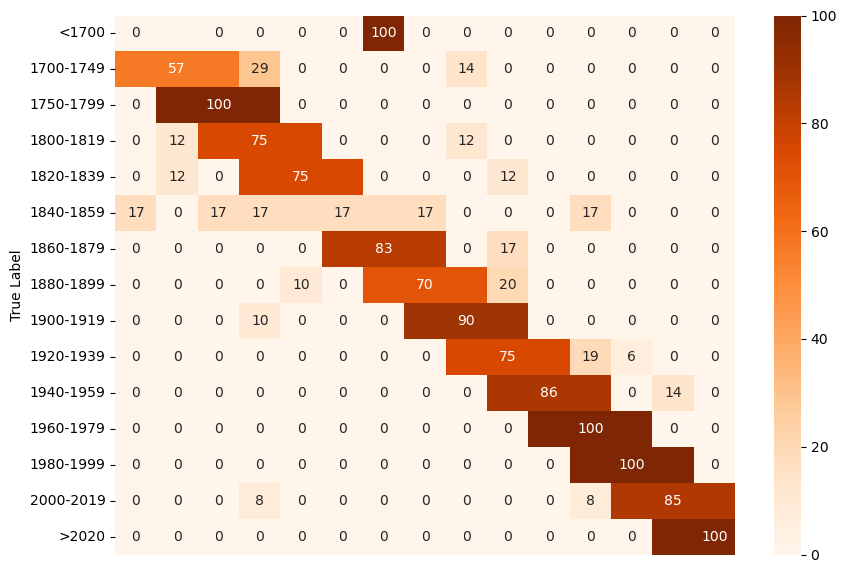}
        }
    \,
    \subfigure{
        \includegraphics[width=0.98\columnwidth]{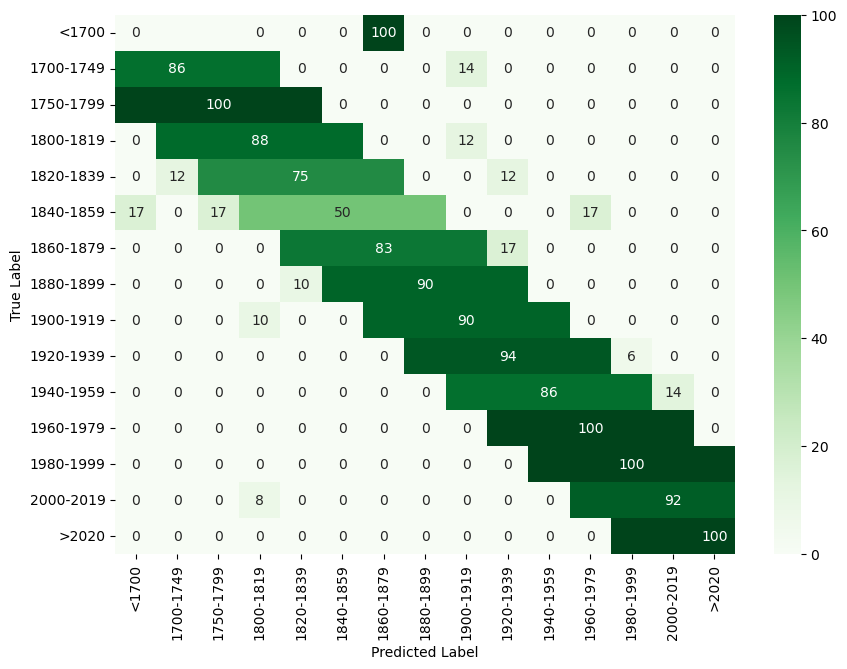}
        }
    \caption{Confusion Matrices. *\textcolor{blue}{Blue matrix} is standard, \textcolor{orange}{Orange matrix} is with adjustment of one adjacent age epoch, and \textcolor{green}{Green matrix} is with adjustment of two adjacent age epochs.}
    \label{fig:cofusion}
\end{figure}

The predictive performance of GPT-4 Vision can be understood more intuitively through the visual analysis of the confusion matrix (shown in Figure \ref{fig:cofusion}. In the standard confusion matrix, we observe that GPT-4 Vision's prediction results show high randomness in certain age epochs, but more prominent performance in the 1800-1819, 1880-1899, 1920-1939, 1960-1979, and 2000-2019 age epochs, which have relatively high recall. This suggests that GPT-4 Vision has a better performance in the prediction of these specific ages.

When a special confusion matrix is introduced - i.e., misclassification of a neighbouring age epoch is also considered as a correct prediction - we find that the vast majority of age epochs are significantly better predicted, with a substantial increase in accuracy. The only less significant predictions are for age epochs $<$1700 and 1840-1859.

Further, when a special confusion matrix is introduced - i.e., misclassification of two neighbouring age epochs is also considered as a correct prediction - we observe a significant improvement in prediction accuracy for almost all categories, including the previously underperforming 1840-1859 age epoch. However, prediction accuracy for age epoch $<$1700 remains a challenge.

In general, these findings reveal the capabilities and limitations of GPT-4 Vision when dealing with the task of classifying building facade images with subtle time-span variations. While the model is able to accurately capture and distinguish age epochs of buildings within a large gaps of years, there is still room for improvement in the predictive accuracy of the model for consecutive age epochs that are stylistically close to each other. 

\section{CONCLUSION}\label{sec:CONCLUSION}
In this study, a training-free classifier based on GPT-4 Vision has been developed to estimate building age epochs from facade images. An enhanced dataset - FI-London combining new facade images with stored building attributes has been proposed. Since GPT by \citet{openai2023gpt4} now provides significant generalisation capability which allows to solve some particular tasks without training data, we designed a reasonable prompt to implement GPT-4 Vision on building age epoch estimation. This is an important and specific task in historical architecture preservation, urban planning and disaster management. Besides traditional evaluation matrices like precision, result and F1 score, we introduced  Mean Absolute Error (MAE) to evaluate the year of classifier missed and visualised several special confusion matrices to show the accuracy in time of the classifier.

Consequently, in 131 tested images from FI-London, the zero-shot classifier predicted 52 correct results and 79 incorrect results including 1 "hallucination" result. Based on human cognition, this "hallucination" result was instead very accurate, due to the rich a-priori knowledge of GPT-4. Based on the analysis of MAE of only 0.85 decade and special confusion matrices, we found that most of the age epochs can be predicted successfully but only approximately, which means they are predicted to be in adjacent but not far off age epochs. However, the construction age is still difficult to predict particularly for old buildings. The reason, based on studies of their input images, may be that these older buildings have been renovated, which causes the models to be easily confused. To sum up, GPT-4 Vision has an ability to estimate the building age epoch in general, but  accurate predictions better than 2 decades and predictions for very old building are still challenging. Furthermore, these results provide valuable insights for further optimisation of the model and highlight the importance of considering architectural style similarities in the building chronology classification task.

In the future, we will continue to explore training-free classifiers in other building attributes such as building occupancy. The framework is not limited to GPT-4, and other VLMs can be considered and compared. Moreover, FI-London will be extended to balance the sample size across age epochs.
\section{Acknowledgement}\label{sec:Acknowledgement}
Z. Zeng and J. M. Goo are supported by the Engineering and Physical Sciences Research Council through an industrial CASE studentship with Ordnance Survey (Grant number EP/W522077/1 and EP/X524840/1).

{
	\begin{spacing}{1.17}
		\normalsize
		\bibliography{references} 
	\end{spacing}
} 

\end{document}